%% file: ACIIDS2016.tex
\documentclass{llncs}
\usepackage{etex}
\usepackage{makeidx}  
\usepackage{lineno,hyperref}
\usepackage[yyyymmdd,hhmmss]{datetime}
\usepackage{color}
\usepackage{multirow}
\usepackage{colortbl}
\usepackage{algorithmic}
\usepackage{algorithm}
\usepackage{amsmath}
\usepackage{amssymb}
\usepackage{makeidx}
\usepackage{graphicx}
\usepackage{multicol} 
\usepackage[bottom]{footmisc}
\usepackage{pst-plot}
\usepackage[normalem]{ulem}
\usepackage{pgfplots}
\usepackage{float}
\usepackage{subcaption}
\usepackage{csvsimple}
\usepackage{longtable}
\usepackage{booktabs}
\usepackage[misc,geometry]{ifsym} 

\pgfplotsset{compat=1.5}
\usetikzlibrary{plotmarks}
\pgfplotsset{every axis/.append style={
                    every x tick label/.append style={font=\tiny},
                    every y tick label/.append style={font=\tiny},
                    every axis label/.append style={font=\small},
                    height=37mm,
                    width=37mm,
                    yshift=-10cm,
                    title style={at={(0.5,0.90)}, font=\normalfont, text height=7mm},
                    x label style={at={(0.5,-0.10)}},
                    mark repeat=5,
            }}
\pgfplotsset{every axis plot/.append style={
                    line width=1.0pt,
            }}

\author{Micha\l{} Spytkowski, \L{}ukasz P. Olech~\Letter, Halina Kwa\'{s}nicka}
\title{Hierarchy of Groups Evaluation Using Different F-score Variants}
\institute{Department of Computational Intelligence, Faculty of Computer Science and Management, Wroc\l{}aw University of Technology, Wroc\l{}aw, Poland\\ Emails: \{michal.spytkowski, lukasz.olech, halina.kwasnicka\}@pwr.edu.pl}
\begin{document}
\mainmatter              
\maketitle
\begin{abstract}
The paper presents a cursory examination of clustering, focusing on a rarely explored field of hierarchy of clusters. Based on this, a short discussion of clustering quality measures is presented and the F-score measure is examined more deeply. As there are no attempts to assess the quality for hierarchies of clusters, three variants of the F-Score based index are presented: classic, hierarchical and partial order. The partial order index is the authors' approach to the subject. Conducted experiments show the properties of the considered measures. In conclusions, the strong and weak sides of each variant are presented.
\end{abstract}
\input{introduction}
\input{fmeasure}
\input{experiments}
\input{conclusions}
\bibliographystyle{splncs03}
\bibliography{mybibfile}
\end{document}

%% file: introduction.tex
\section{Introduction}

Currently we are facing a time in which we are figuratively deluged with data. Thanks to the development and popularity of the Internet, almost every person can take an active part in the creation and distribution of new data. As this mass of data contain potentially useful information, this state presents a new, challenging tasks for scientists. It is no longer feasible to manually obtain this information, and as a result, the development of data mining algorithms has come into focus. Regarding the volume of the data and problems with labelling them, a special interest has turned to unsupervised methods such as cluster analysis, which tries to establish meaningful groups (called \textit{clusters}) in a set of unlabelled data. This approach has been successfully used in many practical applications such as bioinformatics~\cite{Andreopoulos2009}, social media~\cite{Pohl2015}, and audiovisual indexing~\cite{Sevillano2015}. In this paper we make a distinction between three types of clustering which are described below. For the purpose of this paper we only consider situations where a single data point belongs to only one cluster (hard clustering).

The first type is \textbf{flat clustering}, in which data are assigned to independent clusters. This type of clustering does not include relationships between groups. The primary goal of flat clustering is to build a model where data points within any given group are similar between themselves and dissimilar to data points of other groups. In this group two approaches should be mentioned: clusters are generated by iteratively relocating points between subsets, e.g., the \textit{k-means} algorithm~\cite{Hartigan1979,Jain2010} or they are identified as areas of high density of data, e.g., DBSCAN~\cite{Kogan2006}. Most flat clustering methods require to point the number of groups beforehand. The flat clustering methods are not considered in this paper.

Secondly, we consider \textbf{hierarchical clustering} methods. In contrast to flat clustering, they produce a hierarchy of partitions instead of one partition. Each partition differs in the number of clusters. Generally, there are two approaches to hierarchical clustering~\cite{Cimiano2004}: \textit{agglomerative} and \textit{divisive}. 
The agglomerative (\textit{bottom-up} hierarhical clustering) approach starts with each data point in its own cluster and then, iteratively merges two clusters according to a distance function. It ends with all the points in a single cluster, having constructed a hierarchy in the process. The counterpart to the agglomerative approach is the \textit{divisive} approach, also called \textit{deagglomerative} (\textit{top-down} hierarchical clustering). 
The key point in the divisive approach is a function indicating which cluster to split.

The results of hierarchical clustering methods are tree-like structures called \textit{dendrograms}~\cite{Everitt2011}, where nodes represent clusters and the underlying hierarchy shows how the clustering process was performed. As data belong not only to the node it is assigned to, but also to its parents, a dendrogram represents a spectrum of possible clustering results -- from one cluster containing all data to a number of clusters, each containing only one data point. In order to get a specific (flat) partition the dendrogram must be cut. 

In this case the clusters are in a hierarchical relation (partial order), but the data is not. It is because the objects are only assigned to leaves. It remains true that data in a cluster should be maximally similar and that different clusters (at least those containing data) should be maximally dissimilar. Quality measures designed for flat clustering can be successfully applied to hierarchical clustering.

Finally we can consider \textbf{cluster hierarchies}~\cite{ghahramani2010tree,Spytkowski2012}. In this class data can be assigned to any node in the hierarchy of clusters. As in hierarchical clustering there is a relation between groups. \textit{Additionally}, as the data points can be not only in the leaves~\cite{Olech2015}, there is also a hierarchical relation (partial order) between the objects assigned to the clusters. In this case maximum separation between clusters is not always desirable as clusters that are in relation to each other (and thus hold data that is in relations) should be less separated than unrelated clusters. 

Methods generating cluster hierarchies include the Tree Structured Stick Breaking Process~\cite{ghahramani2010tree}, which uses a Markov Chain to explore the space of possible partitions of data or the Bayesian Rose Tree~\cite{blundell2012bayesian} approach, which relies on a deterministic statistical process in order to build the tree. The properties of these clustering approaches appear useful in many fields of information retrieval and processing, especially in documents and images analysis. 
It is however not thoroughly researched if models generated by these methods can be validated in the same way as for more fully explored clustering analysis approaches.

An exhaustive information about clustering can be found in the available literature, for example, in~\cite{MaimonRokach2010,Xu:2005:SCA:2325810.2327433,Madhulatha2012,Kogan2006}.
From the perspective of this paper it is viable to divide existing clusterization techniques based on the characteristics of the result they produce i.e., whether it is a flat or hierarchical structure of clusters. This characteristic underlies our research.

The contribution of this paper is two-fold. Firstly, due to different paradigm of the methods able to build full hierarchies of clusters, we propose to use the name \textbf{\textit{generation of hierarchies of clusters}} (groups) instead of \textit{hierarchical clustering}. In our opinion, such distinction facilitates researchers to note that some new features of hierarchies of clusters are very important from practical point of view~\cite{Spytkowski2012}. The second issue is connected with the evaluation process. The measures useful for evaluation of flat clustering may not be suitable for hierarchies of clusters. In order to at least partially fill this gap, we propose one new external measure, \textbf{\textit{Partial Order F-Score}}. The additional two measures are also studied in experiments. The goal of experiments is to discover suitability of tested measures as well as their advantages and disadvantages. 

The paper is organized as follows. Next section briefly describes problems with quality evaluation of hierarchies of clusters. Third section presents used F-Score based measures. Experiments and their results are described in section 4. Section 5 concludes the paper.

%% file: fmeasure.tex
\section {Measuring Quality of Hierarchies of Clusters}

The above presented problem raises the question if currently used clustering quality measures remain relevant for evaluating hierarchies of groups, and, if not, how can such structures be verified? 
In the case of traditional clustering the core principles~\cite{desgraupes2013clustering} of quality verification can be summarized as data separation between clusters (for internal verification) and data purity within clusters (for external verification). These two approaches can be found at the root of all external and internal clustering measures. These principles fully translate into hierarchical clustering. For internal verification the separation between clusters is still paramount and can be applied on each level at which the hierarchy can be cut. For external verification the lack of data within intermediate clusters means that examining the leaf clusters for purity is all that is left. The data, which is stored in the leaves is not in relation to any data stored in different leaves unless examined level by level.

The perspective changes when we take into account hierarchies of clusters. When every cluster may contain elements the previously defined principles are no longer valid. Data separation becomes problematic as now relationships within clusters need to be considered. While both leaf and sibling clusters should remain maximally separated, the best case scenario for other clusters is not that obvious. As an example, a cluster should be less separated from its descendants and ancestors than it is from completely unrelated clusters. Moreover, for the purity principle, the relations between clusters also come into play. Taken in their classic form, these external measures are completely blind to clusters changing position within the hierarchy.
As the classic measures used in cluster analysis no longer function as intended, the introduction of new measures, or adaptation of existing ones is an explorable field of research. 

\section {F-Score Based Measures}
\textit{F-Score} (also called as F-measure) is commonly used in the field of information retrieval~\cite{Rijsbergen1979}. It is the harmonic mean of the precision and recall measures. F-measure was also adapted for hierarchical clustering~\cite{Mirzaei2008,Olech2015}.
In this paper three different versions of the F-score quality measure are compared to each other. To describe these measures in a clear fashion a number of symbols are introduced:\\
\begin{minipage}[t]{0.5\textwidth}
$X$ -- set of all data points, or objects;\\
$x_i$ -- $i$\textit{-th} data point;\\
$\mathbb{C}$ -- set of all classes $c$;\\
$c_{x_i}$ -- class of object $x_i$;\\
\end{minipage}
\begin{minipage}[t]{0.5\textwidth}
$X_c$ -- set of all objects of class $c$;\\
$\epsilon$ -- specific cluster;\\
$\epsilon_{x_i}$ -- cluster of object $x_i$;\\
$\epsilon\epsilon_i$ -- $i$\textit{-th} child of cluster $\epsilon$ (if exists);\\
$X_\epsilon$ -- set of all objects in cluster $\epsilon$;
\end{minipage}\\

To define the hierarchy of groups a relationship between the ground truth classes must be established:
$C_c = \{c\} \cup \bigcup_{i = 1}^{n}C_{cc_i}$, 
where: $C_c$ -- set containing class $c$ and all its descendant classes; $C_{cc_i}$ -- set containing class $cc_i$ and all its descendant classes; $n$ -- number of children for class $c$.

A relationship between groups in the hierarchy must also be defined: 
$E_\epsilon=\{\epsilon\}\cup\bigcup_{i=1}^{m}E_{\epsilon\epsilon_i}$,
where: $E_\epsilon$ -- set containing node $\epsilon$ and all its descendant nodes; $E_{\epsilon\epsilon_i}$ -- set containing node $\epsilon\epsilon_i$ and all its descendant nodes; $m$ -- number of children for node $\epsilon$.

Using the above defined symbols the set of points belonging to a class (or cluster) and its descendants can be written:
\begin{equation}
X_{C_c} = \bigcup_{c' \in C_c}X_{c'}, \quad\quad\quad X_{E_\epsilon} = \bigcup_{\epsilon' \in E_\epsilon}X_{\epsilon'}.
\end{equation}
F-Score can be viewed as a statistical hypothesis. Thus, it is helpful to define a pair of relations based on pairs of points. In the classic and the partial order approach to F-Score, the statistical hypothesis is calculated based on the total of all possible pairs of different points. Here, the order of the points matters. In this way the condition is based only on the relationship of data points in the ground truth hierarchy of clusters, while the test is based only on the relationship of data points in the tested model. By taking pairs of points and keeping the relationships isolated there is no need to find a correlation between the classes and clusters:
\begin{equation}
G \subseteq X \times X, \quad\quad\quad M \subseteq X \times X,
\end{equation}
where: $G$ -- ground truth relation (condition); $M$ -- model relation (test outcome).

Given this, the four crucial statistical hypothesis values are as follows: 
\begin{equation}
\begin{matrix}
p_t = |\{x_i, x_j \in X: i \neq j \wedge x_i G x_j \wedge x_i M x_j\}|, \\
p_f = |\{x_i, x_j \in X: i \neq j \wedge \neg x_i G x_j \wedge x_i M x_j\}|, \\
n_t = |\{x_i, x_j \in X: i \neq j \wedge \neg x_i G x_j \wedge \neg x_i M x_j\}|, \\
n_f = |\{x_i, x_j \in X: i \neq j \wedge x_i G x_j \wedge \neg x_i M x_j\}|.
\end{matrix}
\end{equation}
where: $p_t$ -- true positive; $p_f$ -- false positive; $n_t$ -- true negative, $n_f$ -- false negative.
\subsection{Classic Clustering F-Score}
In the classic F-Score measure for clustering, we rely on the principle of purity. The perfect clustering result has all of the points in any given cluster belonging to only one class and all of the points within a class belonging to only one cluster. This can be related to the above relations as class and cluster equality: $x_i G x_j \Leftrightarrow c_{x_i} = c_{x_j}$, $x_i M x_j \Leftrightarrow \epsilon_{x_i} = \epsilon_{x_j}$.
Once these relations are defined the classic F-score measure can be calculated as:
\begin{equation}
\label{classicFscore}
F_1 = 2p_t/(2p_t + n_f + p_f).
\end{equation}
\subsection{Hierarchical F-Score}
This measure is an adaptation of F-Score aimed at hierarchical clustering. Initially it was used to measure the quality of clustering in~\cite{Larsen1999}, where the authors tried to create a topic hierarchy with related text documents. The proposed F-measure was used to validate the generated hierarchy as a whole, instead of finding and evaluating a single cut of the dendrogram. Because of this, hierarchical F-Score in its original form can be used in the field of hierarchy of groups. 

For each class $c$ it finds a cluster $\epsilon$ in the hierarchy with the maximal F-measure value:
\begin{equation}
F_c = \max_{\epsilon \in \Theta} \frac{2|X_{E_\epsilon} \cap C_{C_c}|}{|X_{E_\epsilon}| + |X_{C_c}|},
\end{equation}
where: $|X_{E_\epsilon} \cap C_{C_c}|$ -- the number of points belonging to cluster $\epsilon$ and class $c$ as well as their descendants.

When calculating $F_{c}$ we assume that data in descendant clusters and classes belongs to their ancestors, hence $X_{C_c}$ and $X_{E_\epsilon}$ is used. The final quality of a hierarchy is calculated as follows:
\begin{equation}\label{adaptedFmeasure}
F_1 = \frac{\sum_{c \in \mathbb{C}}|X_{C_c}|F_c}{\sum_{c \in \mathbb{C}}|X_{C_c}|}.
\end{equation}
Because this version of the F-measure is a weighted average over all $F_c$, its value is greatly influenced by bigger classes.
In~\cite{Mirzaei2008},~\cite{Olech2015} the calculation of $F_1$ is optimised because data points can belong to only one class. But when working with hierarchies of clusters this is no longer applicable.

\subsection{Partial Order F-Score}
The authors' version of F-Score is a new approach using partial order relation between points that are naturally formed in a hierarchy. This relation can be substituted for equivalence in the $G$ and $M$ relation definition: $x_i G x_j \Leftrightarrow C_{x_i} \subseteq C_{x_j}$, $x_i M x_j \Leftrightarrow E_{x_i} \subseteq E_{x_j}$,
or alternatively: 
$x_i G x_j \Leftrightarrow C_{x_i} \supseteq C_{x_j}$, $x_i M x_j \Leftrightarrow E_{x_i} \supseteq E_{x_j}$.

Both of these are correct as all possible pairs of different points are considered when evaluating a hierarchy. There is a symmetry between the two alternate notations.
By defining the relations between data in this way the measure becomes sensitive to not only the assignment of data from a single class to clusters, but also to the relative position of clusters in the tested model's hierarchy. As such, while it is calculated in the same way as the classic F-measure \textit{(\ref{classicFscore})}, this variant fully makes use of the hierarchy of classes and of existing clusters.

%% file: experiments.tex
\section{Experiments}
To verify the behaviour of the above three measures a series of experiments were conducted on generated data with known properties.
The datasets used during verification were generated using the Tree Structured Stick Breaking Process~\cite{ghahramani2010tree} with eight different sets of parameters,
 each repeated thirty times and averaged. The data has been created by a generator which gives a points' features and class attribute. The values for the parameters used by the generator have been taken from the original publication~\cite{ghahramani2010tree} to showcase a number of different types of hierarchy structures. This selection offers hierarchies of varying width, depth and distributions of the data points among the nodes. The eight datasets used in experiments are as follows:

\begin{minipage}[t]{0.1\textwidth}
Name\\
$\alpha_0$\\
$\lambda$\\
$\gamma$
\end{minipage}
\begin{minipage}[t]{0.1\textwidth}
s00\\
1\\
0.5\\
0.2
\end{minipage}
\begin{minipage}[t]{0.1\textwidth}
s01\\
1\\
1\\
0.2
\end{minipage}
\begin{minipage}[t]{0.1\textwidth}
s02\\
1\\
1\\
1
\end{minipage}
\begin{minipage}[t]{0.1\textwidth}
s03\\
5\\
0.5\\
0.2
\end{minipage}
\begin{minipage}[t]{0.1\textwidth}
s04\\
5\\
1\\
0.2
\end{minipage}
\begin{minipage}[t]{0.1\textwidth}
s05\\
5\\
0.5\\
1
\end{minipage}
\begin{minipage}[t]{0.1\textwidth}
s06\\
25\\
0.5\\
0.2
\end{minipage}
\begin{minipage}[t]{0.1\textwidth}
s07\\
25\\
0.5\\
1
\end{minipage}\\

The three important parameters in the data generation model are $\alpha_0$, $\lambda$ and $\gamma$. These parameters have a series of intuitions about the tree structure associated with them. The alpha function ($\alpha(\epsilon) = \alpha_0\lambda^{|\epsilon|}$) controls the average density of data per level. Considering four cases for the parameters of this function:
\begin{itemize}
\item $\alpha_0$ and $\lambda$ values high (above $1$) lead to extremely sparsely populated deep trees, which is not recommended;
\item $\alpha_0$ and $\lambda$ values low (below $1$) lead to densely populated shallow trees, data is either evenly distributed or grouped closer to the leaves;
\item $\alpha_0$ high, $\lambda$ low leads to shallow trees, data more dense at lower levels of the tree;
\item $\alpha_0$ low, $\lambda$ high leads to deep trees, the data generally grouped close to the top of the tree.
\end{itemize}
\input{random-changes-measure}
When both parameters are close to $1$ the general structure of the tree is hard to predict. The gamma ($\gamma$) parameter controls the average number of children a node can have. Hight values (generally above $1$) lead to trees with more children per node. Lower values (below $1$) lead to trees with less children per node on average. 
All three parameters interact together in the way, that alpha function controls the total data density per level of the hierarchy while $\gamma$ splits the remaining density up between the children.
\subsection{Random Error Introduction Tests}
The first series of experiments carried out involved randomly re-inserting data within the cluster hierarchy. The results are in \figurename~\ref{random-changes-measure}. The probability of a point of data being re-inserted was $\frac{1}{N}$ where $N$ is the total number of data points. 
The re-insertion probability is the same as when generating a model using the TSSB distribution~\cite{ghahramani2010tree}, the distribution parameters remain the same.
The re-insertion process ignores data feature vectors and is based only on the structure of the hierarchy. The re-insertion results in data points changing the cluster they belong to, without changing their class, which the measures should perceive as an error.
%

The results of these experiments show that all measures similarly reacts negatively to the introduction of random errors. The drop rate of the measures appears proportional to the amount of random re-insertions, and none of the measures display any unusual behaviours during this experiment. All of the measures behave as intended in this situation. Since the values of the measures cannot be compared directly to each other it does not matter which variant of F-score achieves higher or lower results.
\subsection{Reduction to a Single Cluster}
In the second experiment all data points were moved into the root node. No random errors were considered during this round of testing. The tests results demonstrate that each of the measures reacts negatively to such extreme under-clustering (Table~\ref{all-in-root-measure}). It is notable however, that the results are no longer as uniform and proportional. Notably the hierarchical F-Score measure stands out when compared with the other two, which show a correlation between their values and descending trend as the mass of data moves generally lower in the generated hierarchies. The value of the hierarchical F-Score however, does not follow this trend.
%
%
%
%
\input{all-in-root-measure}
\subsection{Removing the Cluster Hierarchy}
%
%
The considered measures were tested using data with removed hierarchy. 
All clusters were rendered independently, as if the clustering was flat. 
This reduced the hierarchy to one level and essentially removed the hierarchical relationship between nodes. 
In these experiments errors were also introduced, but the structure of the cluster hierarchy always remained a single-level.
It can be seen in \figurename~\ref{flat-cluster-measure} that the traditional F-measure is oblivious to these changes, as it was expected, because it only considers data points of the same class and cluster.  
However, the other two measures reacted negatively to this flattening.
%
%
\input{flat-cluster-measure}

%% file: random-changes-measure.tex
\noindent
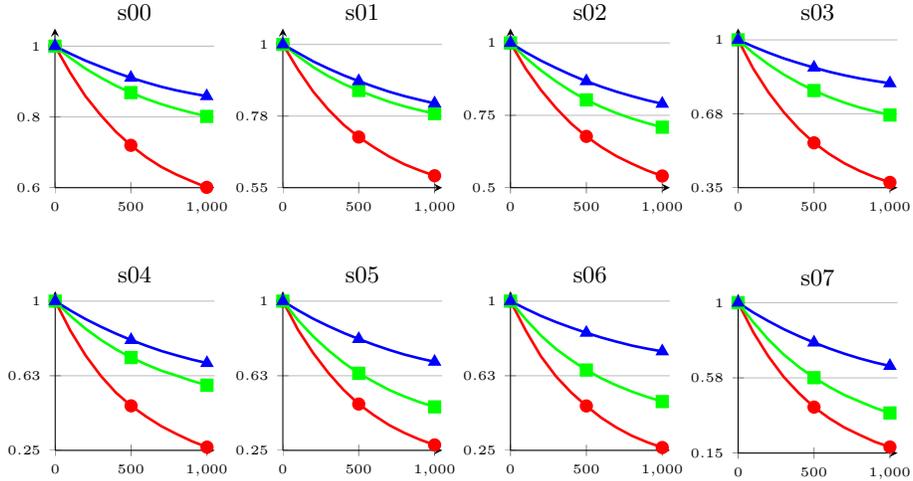
\begin{figure}[t] 
\centering
  \begin{tikzpicture}
    \begin{axis} 
	[
	  title={s00},
      axis lines = left,
	  ymajorgrids = true,
	  ymin = 0.6,
	  ymax = 1.05,
	  xmin = 0,
	  xmax = 1050,
	  ytick={0.6, 0.8, ..., 1.0},
	  xtick={0, 500, 1000},
	  name=s00,
	]
	\addplot
	[
	  color = red,
	  mark = *,
	]
	table
	[
	  col sep = semicolon,
	  x = number-of-random-changes,
	  y = standard-mean,
	]
	{experimentsResults/increasing-random-changes/s00.csv};
	\addplot
	[
	  color = green,
	  mark = square*,
	]
	table
	[
	  col sep = semicolon,
	  x = number-of-random-changes,
	  y = partialOrder-mean,
	]
	{experimentsResults/increasing-random-changes/s00.csv};
	\addplot
	[
	  color = blue,
	  mark = triangle*,
	]
	table
	[
	  col sep = semicolon,
	  x = number-of-random-changes,
	  y = adapted-mean,
	]
	{experimentsResults/increasing-random-changes/s00.csv};
	\end{axis}
    \begin{axis} 
	[
	  title={s01},
      axis lines = left,
	  ymajorgrids = true,
	  ymin = 0.55,
	  ymax = 1.05,
	  xmin = 0,
	  xmax = 1050,
	  ytick={0.55, 0.775, 1.0},
	  xtick={0, 500, 1000},
	  name=s01,
	  at={(150,0)}
	]
	\addplot
	[
	  color = red,
	  mark = *,
	]
	table
	[
	  col sep = semicolon,
	  x = number-of-random-changes,
	  y = standard-mean,
	]
	{experimentsResults/increasing-random-changes/s01.csv};
	\addplot
	[
	  color = green,
	  mark = square*,
	]
	table
	[
	  col sep = semicolon,
	  x = number-of-random-changes,
	  y = partialOrder-mean,
	]
	{experimentsResults/increasing-random-changes/s01.csv};
	\addplot
	[
	  color = blue,
	  mark = triangle*,
	]
	table
	[
	  col sep = semicolon,
	  x = number-of-random-changes,
	  y = adapted-mean,
	]
	{experimentsResults/increasing-random-changes/s01.csv};
	\end{axis}
    \begin{axis} 
	[
	  title={s02},
      axis lines = left,
	  ymajorgrids = true,
	  ymin = 0.5,
	  ymax = 1.05,
	  xmin = 0,
	  xmax = 1050,
	  ytick={0.5, 0.75, 1.0},
	  xtick={0, 500, 1000}, 
	  name=s02,
	  at={(300,0)}
	]
	\addplot
	[
	  color = red,
	  mark = *,
	]
	table
	[
	  col sep = semicolon,
	  x = number-of-random-changes,
	  y = standard-mean,
	]
	{experimentsResults/increasing-random-changes/s02.csv};
	\addplot
	[
	  color = green,
	  mark = square*,
	]
	table
	[
	  col sep = semicolon,
	  x = number-of-random-changes,
	  y = partialOrder-mean,
	]
	{experimentsResults/increasing-random-changes/s02.csv};
	\addplot
	[
	  color = blue,
	  mark = triangle*,
	]
	table
	[
	  col sep = semicolon,
	  x = number-of-random-changes,
	  y = adapted-mean,
	]
	{experimentsResults/increasing-random-changes/s02.csv};
	\end{axis}
    \begin{axis} 
	[
	  title={s03},
      axis lines = left,
	  ymajorgrids = true,
	  ymin = 0.35,
	  ymax = 1.05,
	  xmin = 0,
	  xmax = 1050,
	  ytick={0.35, 0.675, 1.0},
	  xtick={0, 500, 1000},
	  name=s03,
	  at={(450,0)}
	]
	\addplot
	[
	  color = red,
	  mark = *,
	]
	table
	[
	  col sep = semicolon,
	  x = number-of-random-changes,
	  y = standard-mean,
	]
	{experimentsResults/increasing-random-changes/s03.csv};
	\addplot
	[
	  color = green,
	  mark = square*,
	]
	table
	[
	  col sep = semicolon,
	  x = number-of-random-changes,
	  y = partialOrder-mean,
	]
	{experimentsResults/increasing-random-changes/s03.csv};
	\addplot
	[
	  color = blue,
	  mark = triangle*,
	]
	table
	[
	  col sep = semicolon,
	  x = number-of-random-changes,
	  y = adapted-mean,
	]
	{experimentsResults/increasing-random-changes/s03.csv};
	\end{axis}
    \begin{axis} 
	[
	  title={s04},
      axis lines = left,
	  ymajorgrids = true,
	  ymin = 0.25,
	  ymax = 1.05,
	  xmin = 0,
	  xmax = 1050,
	  ytick={0.25, 0.625, 1.0},
	  xtick={0, 500, 1000},
	  name=s04,
	  at={(0,-1320)}
	]
	\addplot
	[
	  color = red,
	  mark = *,
	]
	table
	[
	  col sep = semicolon,
	  x = number-of-random-changes,
	  y = standard-mean,
	]
	{experimentsResults/increasing-random-changes/s04.csv};
	\addplot
	[
	  color = green,
	  mark = square*,
	]
	table
	[
	  col sep = semicolon,
	  x = number-of-random-changes,
	  y = partialOrder-mean,
	]
	{experimentsResults/increasing-random-changes/s04.csv};
	\addplot
	[
	  color = blue,
	  mark = triangle*,
	]
	table
	[
	  col sep = semicolon,
	  x = number-of-random-changes,
	  y = adapted-mean,
	]
	{experimentsResults/increasing-random-changes/s04.csv};
	\end{axis}
    \begin{axis} 
	[
	  title={s05},
      axis lines = left,
	  ymajorgrids = true,
	  ymin = 0.25,
	  ymax = 1.05,
	  xmin = 0,
	  xmax = 1050,
	  ytick={0.25, 0.625, 1.0},
	  xtick={0, 500, 1000},
	  name=s05,
	  at={(150,-1320)}
	]
	\addplot
	[
	  color = red,
	  mark = *,
	]
	table
	[
	  col sep = semicolon,
	  x = number-of-random-changes,
	  y = standard-mean,
	]
	{experimentsResults/increasing-random-changes/s05.csv};
	\addplot
	[
	  color = green,
	  mark = square*,
	]
	table
	[
	  col sep = semicolon,
	  x = number-of-random-changes,
	  y = partialOrder-mean,
	]
	{experimentsResults/increasing-random-changes/s05.csv};
	\addplot
	[
	  color = blue,
	  mark = triangle*,
	]
	table
	[
	  col sep = semicolon,
	  x = number-of-random-changes,
	  y = adapted-mean,
	]
	{experimentsResults/increasing-random-changes/s05.csv};
	\end{axis}
    \begin{axis} 
	[
	  title={s06},
      axis lines = left,
	  ymajorgrids = true,
	  ymin = 0.25,
	  ymax = 1.05,
	  xmin = 0,
	  xmax = 1050,
	  ytick={0.25, 0.625, 1.0},
	  xtick={0, 500, 1000}, 
	  name=s06,
	  at={(300,-1320)}
	]
	\addplot
	[
	  color = red,
	  mark = *,
	]
	table
	[
	  col sep = semicolon,
	  x = number-of-random-changes,
	  y = standard-mean,
	]
	{experimentsResults/increasing-random-changes/s06.csv};
	\addplot
	[
	  color = green,
	  mark = square*,
	]
	table
	[
	  col sep = semicolon,
	  x = number-of-random-changes,
	  y = partialOrder-mean,
	]
	{experimentsResults/increasing-random-changes/s06.csv};
	\addplot
	[
	  color = blue,
	  mark = triangle*,
	]
	table
	[
	  col sep = semicolon,
	  x = number-of-random-changes,
	  y = adapted-mean,
	]
	{experimentsResults/increasing-random-changes/s06.csv};
	\end{axis}
    \begin{axis} 
	[
	  title={s07},
      axis lines = left,
	  ymajorgrids = true,
	  ymin = 0.15,
	  ymax = 1.05,
	  xmin = 0,
	  xmax = 1050,
	  ytick={0.15, 0.575, 1.0},
	  xtick={0, 500, 1000},
	  name=s07,
	  at={(450,-1500)}
	]
	\addplot
	[
	  color = red,
	  mark = *,
	]
	table
	[
	  col sep = semicolon,
	  x = number-of-random-changes,
	  y = standard-mean,
	]
	{experimentsResults/increasing-random-changes/s07.csv};
	\addplot
	[
	  color = green,
	  mark = square*,
	]
	table
	[
	  col sep = semicolon,
	  x = number-of-random-changes,
	  y = partialOrder-mean,
	]
	{experimentsResults/increasing-random-changes/s07.csv};
	\addplot
	[
	  color = blue,
	  mark = triangle*,
	]
	table
	[
	  col sep = semicolon,
	  x = number-of-random-changes,
	  y = adapted-mean,
	]
	{experimentsResults/increasing-random-changes/s07.csv};
	\end{axis}
  \end{tikzpicture}
\caption{Value of F-measures (vertical axis) depending on the number of random re-insertions (horizontal axis), sampled every 100 errors, for each of used datasets: red circles -- Classic F-Score; blue triangles -- Hierarchical F-Score; green squares -- Partial Order F-Score. Note different scale on vertical axes.}
\label{random-changes-measure}
\end{figure}

%% file: all-in-root-measure.tex
\renewcommand\tabcolsep{0pt}
\begin{table}[t] 
	\caption{The average values $\mu$ and standard deviations $\sigma$, all data points are in the root node, without errors introduced.}
	\label{all-in-root-measure}
	\begin{center}
		\begin{tabular}{lcccccc}
			\hline
			\multirow{2}{*}{Set} & \multicolumn{2}{c}{Classic F-score} & \multicolumn{2}{c}{Partial Order F-score} & \multicolumn{2}{c}{Hierarchical F-score} \\
			\cline{2-7} \hspace{6mm} & \hspace{6mm} $\mu$ \hspace{6mm} & \hspace{6mm} $\sigma$ \hspace{6mm} & \hspace{6mm} $\mu$ \hspace{6mm} & \hspace{6mm} $\sigma$ \hspace{7mm} & \hspace{7mm} $\mu$ \hspace{6mm} & \hspace{6mm} $\sigma$ \hspace{6mm} \\
			\hline
			s00 & 0.6456 & 0.1616 & 0.8300 & 0.0857 & 0.8225 & 0.0719 \\
			s01 & 0.5973 & 0.2374 & 0.7977 & 0.1393 & 0.7501 & 0.0966 \\
			s02 & 0.5311 & 0.2600 & 0.7139 & 0.1854 & 0.6970 & 0.1062 \\
			s03 & 0.3729 & 0.1207 & 0.6891 & 0.0915 & 0.7598 & 0.0853 \\
			s04 & 0.1931 & 0.1389 & 0.5737 & 0.1384 & 0.5750 & 0.1089 \\
			s05 & 0.1952 & 0.0953 & 0.4253 & 0.1363 & 0.5680 & 0.0672 \\
			s06 & 0.2031 & 0.0872 & 0.4829 & 0.1368 & 0.6748 & 0.1049 \\
			s07 & 0.0611 & 0.0481 & 0.2528 & 0.0939 & 0.4723 & 0.0918 \\
			\hline
		\end{tabular}
	\end{center}
\end{table}
\renewcommand\tabcolsep{6pt}

%% file: flat-cluster-measure.tex
\noindent
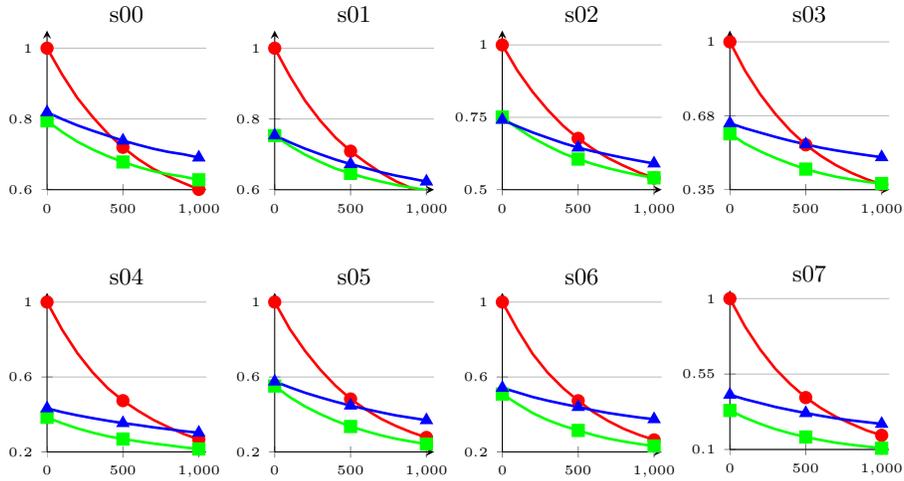
\begin{figure}[t] 
  \begin{tikzpicture}
    \begin{axis} 
	[
	  title={s00},
      axis lines = left,
	  ymajorgrids = true,
	  ymin = 0.6,
	  ymax = 1.05,
	  xmin = 0,
	  xmax = 1050,
	  ytick={0.6, 0.8, 1.0},
	  xtick={0, 500, 1000}, 
	  name=s00,
	]
	\addplot
	[
	  color = red,
	  mark = *,
	]
	table
	[
	  col sep = semicolon,
	  x = number-of-random-changes,
	  y = standard-mean,
	]
	{experimentsResults/flat-cluster-results/s00-flat-clustering.csv};
	\addplot
	[
	  color = green,
	  mark = square*,
	]
	table
	[
	  col sep = semicolon,
	  x = number-of-random-changes,
	  y = partialOrder-mean,
	]
	{experimentsResults/flat-cluster-results/s00-flat-clustering.csv};
	\addplot
	[
	  color = blue,
	  mark = triangle*,
	]
	table
	[
	  col sep = semicolon,
	  x = number-of-random-changes,
	  y = adapted-mean,
	]
	{experimentsResults/flat-cluster-results/s00-flat-clustering.csv};
	\end{axis}
    \begin{axis} 
	[
	  title={s01},
      axis lines = left,
	  ymajorgrids = true,
	  ymin = 0.6,
	  ymax = 1.05,
	  xmin = 0,
	  xmax = 1050,
	  ytick={0.6, 0.8, 1.0},
	  xtick={0, 500, 1000}, 
	  name=s01,
	  at={(150,0)}
	]
	\addplot
	[
	  color = red,
	  mark = *,
	]
	table
	[
	  col sep = semicolon,
	  x = number-of-random-changes,
	  y = standard-mean,
	]
	{experimentsResults/flat-cluster-results/s01-flat-clustering.csv};
	\addplot
	[
	  color = green,
	  mark = square*,
	]
	table
	[
	  col sep = semicolon,
	  x = number-of-random-changes,
	  y = partialOrder-mean,
	]
	{experimentsResults/flat-cluster-results/s01-flat-clustering.csv};
	\addplot
	[
	  color = blue,
	  mark = triangle*,
	]
	table
	[
	  col sep = semicolon,
	  x = number-of-random-changes,
	  y = adapted-mean,
	]
	{experimentsResults/flat-cluster-results/s01-flat-clustering.csv};
	\end{axis}
    \begin{axis} 
	[
	  title={s02},
      axis lines = left,
	  ymajorgrids = true,
	  ymin = 0.5,
	  ymax = 1.05,
	  xmin = 0,
	  xmax = 1050,
	  ytick={0.5, 0.75, 1.0},
	  xtick={0, 500, 1000},
	  name=s02,
	  at={(300,0)}
	]
	\addplot
	[
	  color = red,
	  mark = *,
	]
	table
	[
	  col sep = semicolon,
	  x = number-of-random-changes,
	  y = standard-mean,
	]
	{experimentsResults/flat-cluster-results/s02-flat-clustering.csv};
	\addplot
	[
	  color = green,
	  mark = square*,
	]
	table
	[
	  col sep = semicolon,
	  x = number-of-random-changes,
	  y = partialOrder-mean,
	]
	{experimentsResults/flat-cluster-results/s02-flat-clustering.csv};
	\addplot
	[
	  color = blue,
	  mark = triangle*,
	]
	table
	[
	  col sep = semicolon,
	  x = number-of-random-changes,
	  y = adapted-mean,
	]
	{experimentsResults/flat-cluster-results/s02-flat-clustering.csv};
	\end{axis}
    \begin{axis} 
	[
	  title={s03},
      axis lines = left,
	  ymajorgrids = true,
	  ymin = 0.35,
	  ymax = 1.05,
	  xmin = 0,
	  xmax = 1050,
	  ytick={0.35, 0.675, 1.0},
	  xtick={0, 500, 1000}, 
	  name=s03,
	  at={(450,0)} 
	]
	\addplot
	[
	  color = red,
	  mark = *,
	]
	table
	[
	  col sep = semicolon,
	  x = number-of-random-changes,
	  y = standard-mean,
	]
	{experimentsResults/flat-cluster-results/s03-flat-clustering.csv};
	\addplot
	[
	  color = green,
	  mark = square*,
	]
	table
	[
	  col sep = semicolon,
	  x = number-of-random-changes,
	  y = partialOrder-mean,
	]
	{experimentsResults/flat-cluster-results/s03-flat-clustering.csv};
	\addplot
	[
	  color = blue,
	  mark = triangle*,
	]
	table
	[
	  col sep = semicolon,
	  x = number-of-random-changes,
	  y = adapted-mean,
	]
	{experimentsResults/flat-cluster-results/s03-flat-clustering.csv};
	\end{axis}
    \begin{axis} 
	[
	  title={s04},
      axis lines = left,
	  ymajorgrids = true,
	  ymin = 0.2,
	  ymax = 1.05,
	  xmin = 0,
	  xmax = 1050,
	  ytick={0.2, 0.6, 1.0},
	  xtick={0, 500, 1000}, 
	  name=s04,
	  at={(0,-1400)}
	]
	\addplot
	[
	  color = red,
	  mark = *,
	]
	table
	[
	  col sep = semicolon,
	  x = number-of-random-changes,
	  y = standard-mean,
	]
	{experimentsResults/flat-cluster-results/s04-flat-clustering.csv};
	\addplot
	[
	  color = green,
	  mark = square*,
	]
	table
	[
	  col sep = semicolon,
	  x = number-of-random-changes,
	  y = partialOrder-mean,
	]
	{experimentsResults/flat-cluster-results/s04-flat-clustering.csv};
	\addplot
	[
	  color = blue,
	  mark = triangle*,
	]
	table
	[
	  col sep = semicolon,
	  x = number-of-random-changes,
	  y = adapted-mean,
	]
	{experimentsResults/flat-cluster-results/s04-flat-clustering.csv};
	\end{axis}
    \begin{axis} 
	[
	  title={s05},
      axis lines = left,
	  ymajorgrids = true,
	  ymin = 0.2,
	  ymax = 1.05,
	  xmin = 0,
	  xmax = 1050,
	  ytick={0.2, 0.6, 1.0},
	  xtick={0, 500, 1000}, 
	  name=s05,
	  at={(150,-1400)}
	]
	\addplot
	[
	  color = red,
	  mark = *,
	]
	table
	[
	  col sep = semicolon,
	  x = number-of-random-changes,
	  y = standard-mean,
	]
	{experimentsResults/flat-cluster-results/s05-flat-clustering.csv};
	\addplot
	[
	  color = green,
	  mark = square*,
	]
	table
	[
	  col sep = semicolon,
	  x = number-of-random-changes,
	  y = partialOrder-mean,
	]
	{experimentsResults/flat-cluster-results/s05-flat-clustering.csv};
	\addplot
	[
	  color = blue,
	  mark = triangle*,
	]
	table
	[
	  col sep = semicolon,
	  x = number-of-random-changes,
	  y = adapted-mean,
	]
	{experimentsResults/flat-cluster-results/s05-flat-clustering.csv};
	\end{axis}
    \begin{axis} 
	[
	  title={s06},
      axis lines = left,
	  ymajorgrids = true,
	  ymin = 0.2,
	  ymax = 1.05,
	  xmin = 0,
	  xmax = 1050,
	  ytick={0.2, 0.6, 1.0},
	  xtick={0, 500, 1000}, 
	  name=s06,
	  at={(300,-1400)}
	]
	\addplot
	[
	  color = red,
	  mark = *,
	]
	table
	[
	  col sep = semicolon,
	  x = number-of-random-changes,
	  y = standard-mean,
	]
	{experimentsResults/flat-cluster-results/s06-flat-clustering.csv};
	\addplot
	[
	  color = green,
	  mark = square*,
	]
	table
	[
	  col sep = semicolon,
	  x = number-of-random-changes,
	  y = partialOrder-mean,
	]
	{experimentsResults/flat-cluster-results/s06-flat-clustering.csv};
	\addplot
	[
	  color = blue,
	  mark = triangle*,
	]
	table
	[
	  col sep = semicolon,
	  x = number-of-random-changes,
	  y = adapted-mean,
	]
	{experimentsResults/flat-cluster-results/s06-flat-clustering.csv};
	\end{axis}
    \begin{axis} 
	[
	  title={s07},
      axis lines = left,
	  ymajorgrids = true,
	  ymin = 0.1,
	  ymax = 1.05,
	  xmin = 0,
	  xmax = 1050,
	  ytick={0.1, 0.55, 1.0},
	  xtick={0, 500, 1000}, 
	  name=s07,
	  at={(450,-1550)}
	]
	\addplot
	[
	  color = red,
	  mark = *,
	]
	table
	[
	  col sep = semicolon,
	  x = number-of-random-changes,
	  y = standard-mean,
	]
	{experimentsResults/flat-cluster-results/s07-flat-clustering.csv};
	\addplot
	[
	  color = green,
	  mark = square*,
	]
	table
	[
	  col sep = semicolon,
	  x = number-of-random-changes,
	  y = partialOrder-mean,
	]
	{experimentsResults/flat-cluster-results/s07-flat-clustering.csv};
	\addplot
	[
	  color = blue,
	  mark = triangle*,
	]
	table
	[
	  col sep = semicolon,
	  x = number-of-random-changes,
	  y = adapted-mean,
	]
	{experimentsResults/flat-cluster-results/s07-flat-clustering.csv};
	\end{axis}
  \end{tikzpicture}
\caption{Value of F-measures (vertical axis) depending on the number of random re-insertions (horizontal axis), sampled every 100 errors, for data with removed hierarchy: red circles -- Classic F-Score; blue triangles -- Hierarchical F-Score; green squares -- Partial Order F-Score. Note different scale on vertical axes.}
\label{flat-cluster-measure}
\end{figure}

%% file: conclusions.tex
\section{Conclusions}
Based on the results of the conducted experiments it is possible to comment the general behaviour of the three measures. Each of them possesses strengths and weaknesses. Ultimately the choice depends on the problem being evaluated.

\textbf{Classic F-Score} is based strictly on hypothesis tests and reflects relations found in the flat and hierarchical clustering. This measure can reach both the maximum and minimum value and is simple to calculate. But the important weakness of classic F-Score is that it does not work properly for hierarchies of clusters and it notices fewer types of errors than the other measures.

\textbf{Hierarchical F-Score} reflects relations found in many types of structures (flat clustering, hierarchies, forests of hierarchies),
in some cases it can be optimised to work more efficiently~\cite{Mirzaei2008,Olech2015}.
Because it is a weighted sum, it focuses more on the numerous classes, which can be potentially useful. Besides the above strengths, we should mention some weaknesses. It is not based on hypothesis tests, cannot possibly reach its minimal value, points on lower levels of the hierarchy contribute to the final result with a higher weight than points higher up. Unoptimised version requires more complex calculation.

\textbf{Partial Order F-Score} reflects relations found in many types of structures (including flat clustering, hierarchies, and forests of hierarchies), is capable of reaching both the maximum and minimum value, and is based strictly on hypothesis tests. With points assigned only to leaf nodes, it is indistinguishable from the classic F-Score, therefore for flat clusters it can play the role of classic F-Score. It can be optimised to work as fast as classic F-score. 
However, when unoptimised, is more complex to calculate than classic F-Score, which should be mentioned as its weakness.

The paper offers some interesting insights into the adaptation of the F-Score measure to work with hierarchies of clusters. However, there are other common clustering measures that can be adapted in similar ways. Further research is required in order to judge the feasibility of other quality indices. We are continuing research on the generation of hierarchies of clusters as well as on the external and internal measures suitable to validate and compare their results.

\textit{Acknowledgements}: The research was supported by the European Commission under the 7th Framework Programme, Coordination and Support Action, Grant Agreement Number 316097, ENGINE -- European research centre of Network intelliGence for INnovation Enhancement (http://engine.pwr.wroc.pl/).

%% file: ACIIDS2016.bbl
\begin{thebibliography}{10}
\providecommand{\url}[1]{\texttt{#1}}
\providecommand{\urlprefix}{URL }

\bibitem{Andreopoulos2009}
Andreopoulos, B., An, A., Wang, X., Schroeder, M.: A roadmap of clustering
  algorithms: finding a match for a biomedical application. Briefings in
  Bioinformatics  10(3),  297--314 (2009)

\bibitem{blundell2012bayesian}
Blundell, C., Teh, Y.W., Heller, K.A.: Bayesian rose trees. arXiv preprint
  arXiv:1203.3468  (2012)

\bibitem{Cimiano2004}
Cimiano, P., Hotho, A., Staab, S.: Comparing conceptual, divise and
  agglomerative clustering for learning taxonomies from text. In:
  de~M{\'{a}}ntaras, R.L., Saitta, L. (eds.) Proc. of the 16th Eureopean Conf.
  on AI, Spain, 2004. pp. 435--439. {IOS} Press (2004)

\bibitem{desgraupes2013clustering}
Desgraupes, B.: Clustering indices  (2013),
  \url{https://cran.r-project.org/web/packages/clusterCrit/vignettes/clusterCrit.pdf}

\bibitem{Everitt2011}
Everitt, B.S., Landau, S., Leese, M., Stahl, D.: Cluster Analysis. John Wiley
  and Sons, Ltd (2011)

\bibitem{ghahramani2010tree}
Ghahramani, Z., Jordan, M.I., Adams, R.P.: Tree-structured stick breaking for
  hierarchical data. In: NIPS. pp. 19--27 (2010)

\bibitem{Hartigan1979}
Hartigan, J.A., Wong, M.A.: Algorithm as 136: A k-means clustering algorithm.
  Applied Statistics  28(1),  100--108 (1979)

\bibitem{Jain2010}
Jain, A.K.: Data clustering: 50 years beyond k-means. Pattern Recognition
  Letters  31(8),  651--666 (2010)

\bibitem{Kogan2006}
Kogan, J., Nicholas, C.K., Teboulle, M.: Grouping multidimensional data recent
  advances in clustering (2006)

\bibitem{Larsen1999}
Larsen, B., Aone, C.: Fast and effective text mining using linear-time document
  clustering. In: Fayyad, U.M., Chaudhuri, S., Madigan, D. (eds.) Proc. of the
  5th {ACM} {SIGKDD} Inter. Conf. on Knowledge Discovery and Data Mining, USA.
  pp. 16--22. {ACM} (1999)

\bibitem{Madhulatha2012}
Madhulatha, T.S.: An overview on clustering methods. CoRR  abs/1205.1117 (2012)

\bibitem{Mirzaei2008}
Mirzaei, A., Rahmati, M., Ahmadi, M.: A new method for hierarchical clustering
  combination. Intell. Data Anal.  12(6),  549--571 (Dec 2008)

\bibitem{MaimonRokach2010}
Oded, M., Lior, R. (eds.): Data Mining and Knowledge Discovery Handbook.
  Springer (2010)

\bibitem{Olech2015}
Olech, L.P., Paradowski, M.: Hierarchical gaussian mixture model with objects
  attached to terminal and non-terminal dendrogram nodes. In: 9th International
  Conf. on Computer Recognition Systems, Poland, 2015 (2015)

\bibitem{Pohl2015}
Pohl, D., Bouchachia, A., Hellwagner, H.: Social media for crisis management:
  clustering approaches for sub-event detection. Multimedia Tools and
  Applications  74(11),  3901--3932 (2015)

\bibitem{Rijsbergen1979}
van Rijsbergen, C.J.: Information Retrieval. Butterworth (1979)

\bibitem{Sevillano2015}
Sevillano, X., Valero, X., AlĂ­as, F.: Look, listen and find: A purely
  audiovisual approach to online videos geotagging. Information Sciences  295,
  558 -- 572 (2015)

\bibitem{Spytkowski2012}
Spytkowski, M., Kwasnicka, H.: Hierarchical clustering through bayesian
  inference. In: Nguyen, N.T., Hoang, K., Jedrzejowicz, P. (eds.) ICCCI (1).
  Lecture Notes in Computer Science, vol. 7653, pp. 515--524. Springer (2012)

\bibitem{Xu:2005:SCA:2325810.2327433}
Xu, R., Wunsch, II, D.: Survey of clustering algorithms. Trans. Neur. Netw.
  16(3),  645--678 (May 2005)

\end{thebibliography}
